\documentclass[conference]{IEEEtran}
\IEEEoverridecommandlockouts
\usepackage{cite}
\usepackage{amsmath,amssymb,amsfonts}
\usepackage{algorithmic}
\usepackage{graphicx}
\usepackage{textcomp}
\usepackage{xcolor}
\usepackage{amsmath}

\def\BibTeX{{\rm B\kern-.05em{\sc i\kern-.025em b}\kern-.08em
    T\kern-.1667em\lower.7ex\hbox{E}\kern-.125emX}}
\begin{document}

\title{Multi-scale Feature Imitation for Unsupervised Anomaly Localization}

\author{\IEEEauthorblockN{1\textsuperscript{st} Chao Hu}
\IEEEauthorblockA{\textit{AI Lab} \\
\textit{Unicom (Shanghai) Industry Internet Co., Ltd.}\\
Shanghai, China \\
huchao.000@gmail.com}
\and
\IEEEauthorblockN{2\textsuperscript{nd} Shengxin Lai}
\IEEEauthorblockA{\textit{AI Lab} \\
\textit{Unicom (Shanghai) Industry Internet Co., Ltd.}\\
Shanghai, China \\
laishengxin18@gmail.com}
}

\maketitle

\begin{abstract}
The unsupervised anomaly localization task faces the challenge of missing anomaly sample training, detecting multiple types of anomalies, and dealing with the proportion of the area of multiple anomalies. A separate teacher-student feature imitation network structure and a multi-scale processing strategy combining an image and feature pyramid are proposed to solve these problems. A network module importance search method based on gradient descent optimization is proposed to simplify the network structure. The experimental results show that the proposed algorithm performs better than the feature modeling anomaly localization method on the real industrial product detection dataset in the same period. The multi-scale strategy can effectively improve the effect compared with the benchmark method.
\end{abstract}


\section{Introduction}
Anomaly detection is a potentially important task in some computer vision information processing systems, such as behavior monitoring \cite{ref1,ref2}, automated manufacturing process inspection \cite{ref3, ref4, ref5}, and medical diagnosis \cite{ref6} and disease monitoring. The difficulty of anomaly detection tasks lies in the scarcity and unknown of anomaly patterns, and it is challenging to capture anomalous samples. In this case, getting enough anomalous samples as training data is difficult. An unsupervised learning paradigm is more suitable for anomaly detection tasks: training the model and establishing decision boundaries using only normal samples.

\begin{figure*}
    \centering
    \includegraphics[width=0.9\textwidth]{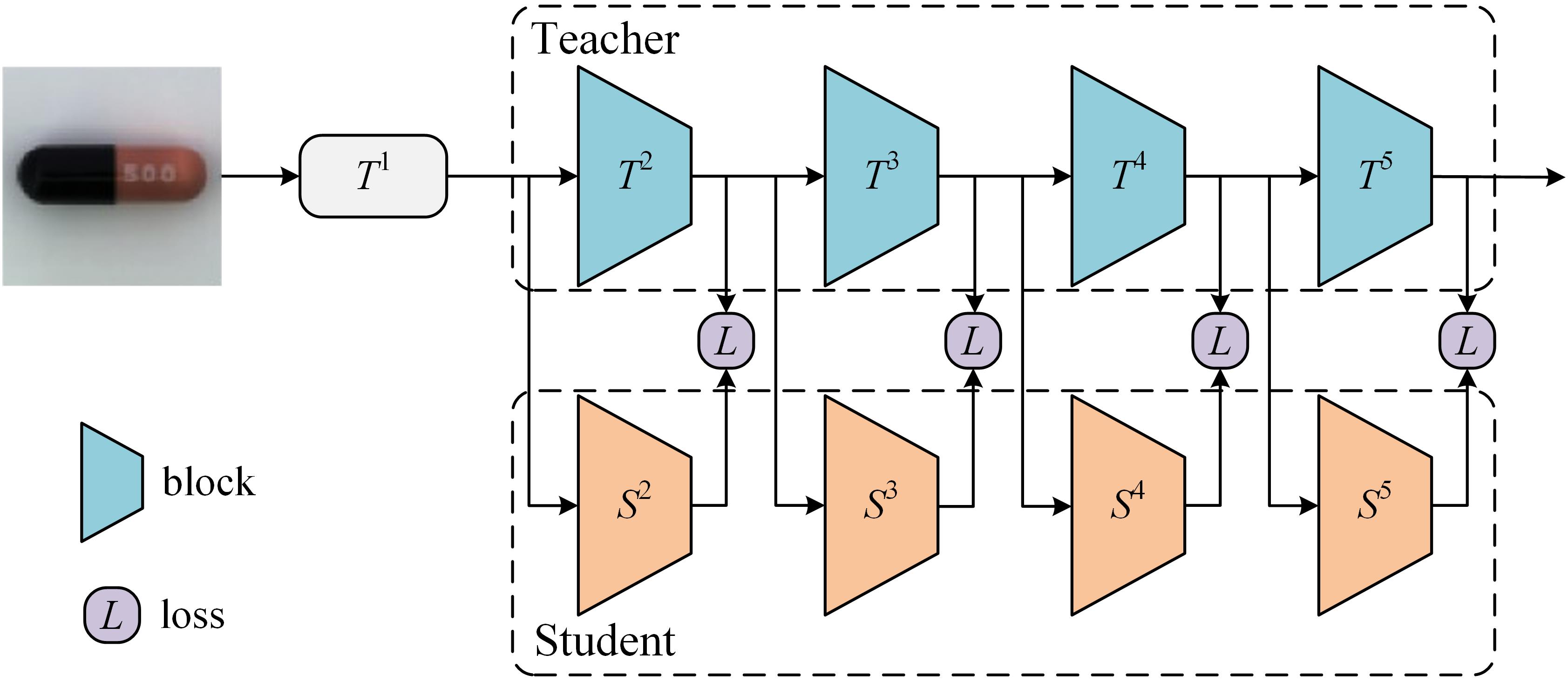}
    \caption{\centering{Overall network structure of teacher-student imitation}.}
    \label{fig1}
\end{figure*}

There are three main methods of image-level anomaly detection. (1) Methods based on reconstruction, such as using autoencoders \cite{ref7}, variational autoencoders \cite{ref8}, generative adversarial networks \cite{ref9}, adding memory modules \cite{ref10}, etc., to identify abnormal image data by comparing the differences between the reconstructed image and the original image; (2) Based on the distribution method, construct the probability distribution of normal data, and identify data with low probability density values as abnormalities, such as the methods of using Gaussian mixture models DAGMM \cite{ref11} and ADGAN \cite{ref12}; (3) Based on the classification method, build a data feature training classifier for discrimination, such as a classification support vector machine (OC-SVM) \cite{ref13}, use the support vector data description of deep network features (Deep-SVDD) \cite{ref14}, combined with the fully supervised classification task to train a classifier \cite{ref15} etc. The Patch SVDD \cite{ref16} method extends SVDD to the task of anomaly localization based on image blocks.

For pixel-level anomaly detection, the generation-based method uses the generative model to generate the image closest to the input image and compare it with the input image at the pixel level. For example, Baur \cite{ref17} and others use self-encoders to compare the reconstructed image with the input image; Schlegl et al. \cite{ref18} use generative adversarial networks to find the reconstructed image closest to the input image from hidden space. Some recent work combines class activation graphs for anomalous image segmentation. Venkataramanan et al. \cite{ref19} proposed a combined variational autoencoder and attention method (CAVGA); Salehi et al. \cite{ref20} combined knowledge distillation models and various class activation graphs. Many methods utilize highly expressive pre-trained networks to construct normal data pattern distributions. Bergmann et al. \cite{ref21} transferred the pre-trained network knowledge to the image block describer and used feature representation differences to identify abnormal image blocks. Defard et al. \cite{ref22} used the multi-intermediate layer feature stitching of the pre-trained network as data feature description and Gaussian distribution to describe the normal data feature distribution and Mahalanobis distance to measure the degree of anomaly. Cohen et al. \cite{ref23} construct feature pyramids and use k-nearest neighbor clustering to identify anomalous data.

Some work utilizes feature networks trained on sizeable natural image datasets. Bergmann et al. \cite{ref21} propose an anomaly detection framework based on a teacher-student structure. They migrated highly expressive features from robust pre-trained networks to teacher networks through the knowledge distillation process, used the features of teacher networks as regression targets for student networks, and then used the representation error of student networks and prediction uncertainty between students as the basis for anomaly discrimination. This method uses local overlapping image block feature embedding to obtain pixel-level discrimination, resulting in low inference efficiency. This paper adopts the feature map comparison method \cite{ref24} from the image block modeling method of this method. Feature maps are generated naturally by the neural network, so feature maps from the entire input image can be used directly for discrimination. Due to this feature, the proposed method does not need to train the image block descriptor and can obtain a pixel-level anomalous discriminant score map through forward inference, which significantly improves the inference efficiency.

In this paper, an intermediate feature regression and comparison method are proposed to split the student network into sub-blocks, the output of each sub-block mimics the characteristics of the corresponding position of the teacher network, and the input of the sub-blocks comes from the features of the previous layer of teacher networks, which has a more stable training performance. In addition, this paper proposes a multi-scale processing strategy, combining image pyramid operation and maintaining multi-feature layer learning, that is, using multiple input size images for training and testing. However, it has been observed that training a model on input image scaling can severely damage model performance. Therefore, this paper maps each input size to different student block groups and vertically forms a multi-level student network structure. In this case, more input dimensions mean more computational overhead. Therefore, this paper designs a side task to set a weight for each student network block and optimize these weights using gradient descent based on an anomaly data validation set. This process and the main task of model training are independent of each other, so network parameters are not affected by anomalous data. In this way, more efficient multi-scale processing is possible.

\section{Approach}
This paper's primary method for anomaly localization is knowledge distillation feature regression and comparison. The normal sample is input to the teacher and student network. The mid features of the student network imitate the corresponding features of the teacher network, so it can be considered that the abnormal pattern that did not occur in the training process is more different in the middle features of the student network and the middle features of the teacher network. We represent this difference using a distance metric. Therefore, the distance between student network features and teacher network features can be used as the basis for judgment. The student subgroup structure proposed in this paper avoids the interaction of different layers of supervision information, makes the training process more stable, and improves the effect of features imitation. In addition, this structure is flexible for the multi-scale strategy used in this paper, which can reduce the computational burden and avoid retraining the network. This section details the network structure, multi-scale processing strategy, and module weight search auxiliary tasks.

\subsection{Teacher-student Feature Imitation Framework}
Unlike the teacher-student structure used in other work, this paper designs a separate student network block group. The overall structure is shown in Fig. \ref{fig1}. A teacher network is a complete network structure in which middle features serve as the learning objectives of the student network. Student networks are separate network modules, each having the same module structure as the corresponding teacher network. The previous layer of the teacher network passes through this module to generate the current layer of student features. Then, the loss function is constructed based on the gap between the characteristics of students and teachers in the current layer. According to the idea of knowledge distillation work \cite{ref25}, this paper uses multi-layer middle features for guidance and learning. On the one hand, different hidden layers in neural networks encode features of different levels of abstraction; On the other hand, different feature layers have different levels of receptive fields, which is very important for inferring anomalous regions with different area proportions.

\begin{figure}[!t]
	\centering
     \includegraphics[width=0.45\textwidth]{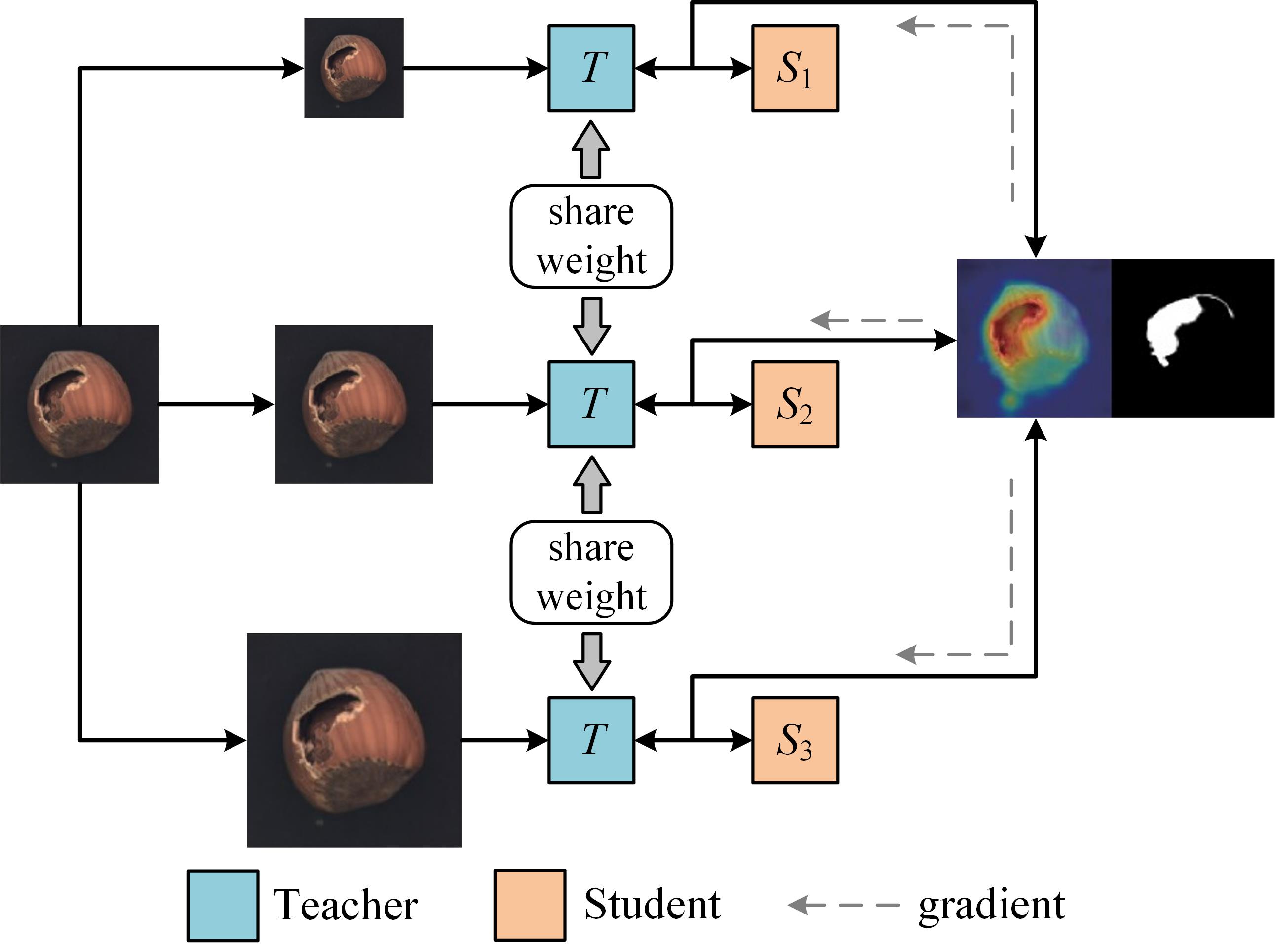}
	\caption{\centering{Multi-scale teacher-student network structure and scale weight search}.}
	\label{fig2}
\end{figure}

\subsubsection{Training Process}
Given a training set containing only normal samples, the training goal is to minimize the distance between the student and teacher network features on the normal samples. As a guide, the teacher network is a robust pre-training network whose parameters are fixed throughout the training and testing process. As shown in Fig. \ref{fig1}, teachers are represented as different tiers of $T^l$ and students as $S^l$. We use the middle feature map $\{f^l\}^L_{l=2}$ for feature matching, and the feature map size is $c_l \times h_l \times w_l$, and the $L$-layer features are used together. The output features of the teacher network are represented as $f^l_t=T^l(f_t^{l-1})$, $f_t^1=T^1(I_k)$, and $I_k$ as the input image. The output features of the student network are expressed as $f^l_s=S^l(f_t^{l-1})$. All feature maps are $L_2$ normalized at each pixel location, and the learning goal is to minimize the distance between the teacher's and student's features.
\begin{equation}
\label{equation:1}
\mathcal{L}^l=\frac{1}{h_lw_l}\sum_{p}{\| \widehat{f}^l_{t(p)}-\widehat{f}^l_{s(p)} \| }^2_2 
\end{equation}
\begin{equation}
\label{equation:2}
\widehat{f}_{(p)}=\frac{f_{(p)}}{\|f_{(p)}\|_2}
\end{equation}

Where $p$ represents a location in the feature map, standardizing all features makes training convergence more stable. For multi-level feature learning, the overall loss is:
\begin{equation}
\label{equation:3}
\mathcal{L}=\sum_{L}^{l=2}\mathcal{L}^l
\end{equation}

\subsubsection{Inference Process}
For abnormal localization, it is hoped that each input image pixel can predict a score and that the high discriminant score is abnormal. The reasoning standard is the same as the training standard. For all feature layers, the distance between teacher features and student features is calculated pixel by pixel to obtain the score map of this feature layer. Here all score maps are upsampled to the input image size, and the average of all score maps is taken as the final discriminative score map.
\begin{equation}
\label{equation:4}
\phi(I)=\frac{1}{L}\sum_{l=2}^{L}unsample(\phi^l(I))
\end{equation}
\begin{equation}
\label{equation:5}
\phi^l_{(p)}(I)=\|\widehat{f}^l_{t(p)}-\widehat{f}^l_{s(p)}\|^2_2
\end{equation}
\begin{equation}
\label{equation:6}
\widehat{f}_{(p)}=\frac{f(p)}{\|f_{(p)}\|_2}
\end{equation}

The fraction of all feature locations is the square distance between the standardized vectors, and the value range is $[0, 4]$. Multiply the fraction plot by a scale factor of 0.25 so that its value is limited to $[0, 1]$.

With all student modules independent of each other,  different student modules can be disassembled and combined at will. This structure allows us to search for the optimal combination of student modules without retraining the model, significantly saving module search time. At the same time, dismantling inefficient student modules can increase the inference efficiency of the model and save memory expenses. Therefore, based on this structure, this paper proposes a flexible multi-scale processing strategy and module importance search strategy.

\subsubsection{Multi-scale Processing Strategy}
The above proposes a teacher-student framework and a multi-middle learning approach. Different middle layers of neural networks encode features at different levels, such as shallow coding details or structural features, while deep encoding semantic-level features. The multi-level receptive fields of the feature layer realize the information aggregation of image regions of different sizes. Using feature pyramids can meet the needs of multi-scale detection to a certain extent. Nevertheless, this fixed scale setting does not satisfy the many categories and types of detected defects. Therefore, this paper incorporates image pyramids into existing frameworks.

An intuitive way to handle this is to set the input images to different sizes and input them into the same network for training and testing, a data augmentation method widely used in discriminant models. However, this approach is not suitable for anomaly detection tasks because the augmented transformation of the input data will enhance the generalization ability of the model, thereby generalizing the pattern that is normal to the anomaly category and cannot clearly define the behavior of the anomaly class in the model, thus breaking the boundary between a normal class and anomalous class. This article solves this problem by building different models for different input sizes. For each additional input scale layer, a group of student module groups of $S_i=\{S^l|l=2, 3, \dots, L\}$ is added to the network. All student modules are separated, so their intermediate representations do not interfere. The teacher network is a fixed pre-training network that all student networks can share. The overall structure is shown in Fig. \ref{fig2}. Student module networks can be trained independently. The score maps calculated from the student module group hierarchy corresponding to the image scale and the multi-layer feature maps within the level are interpolated to reference size. The final discriminant score map is obtained by averaging.

\subsubsection{Scale Weight Search}
Although the multi-scale approach can lead to better performance, adding model units will undoubtedly add additional computational overhead. To overcome this limitation, this paper assigns a weight to each student block, sets up a validation set consisting of anomalous images and their labels, and determines these weights by the performance of the validation set. After determining the weight of each student module, we regard the weight size as the importance of the student module, and the low-weight module can be discarded directly.

Specifically, this paper uses gradient descent for weight determination. The score plots for each student block are combined in a weighted form. Considering that anomalies are position-independent, so all pixels in the image share the same weight, the final score plot $\phi$ is:
\begin{equation}
\label{equation:7}
\phi=\sum_{i=1}^{N}\omega _i\phi_i
\end{equation}
\begin{equation}
\label{equation:8}
w_i=\frac{e^{\widehat{\omega}_{i}}}{\sum_{j}e^{\widehat{\omega}_{j}}}
\end{equation}

$N$ represents the number of student modules, and $\omega_i$ represents the weight assigned to the fraction plot $F_i$, calculated by normalizing the module parameter $\widehat{\omega}_{i}$. During the validation phase, the parameter $\widehat{\omega}_i$ is updated using gradient descent by searching for the validation set.
\begin{equation}
\label{equation:9}
\widehat{\omega}_{i}^{m+1}=\widehat{\omega}_{i}^{m}-\xi\nabla_{\widehat{\omega}_{i}}\mathcal{L}_{val}
\end{equation}

\noindent where $m$ is the number of iterations, $\xi$ is the learning rate, and $\mathcal{L}_{val}$ represents the cross-entropy loss.
\begin{equation}
\label{equation:10}
\mathcal{L}_{val}=CE(\sum_{i=1}^{N}\omega_{i}\phi_i, gt)
\end{equation}

After the network has been trained to a certain extent, fix the parameters of all student network modules, set the module weights to learnable modules, and optimize them. The validation phase optimizes the module weights, independent of the training phase of optimizing network parameters. After the module weight is determined, we regard the high-weight module as more critical, so we keep the $k$ modules with the highest weight and discard the remaining modules directly. Because student modules are separate, modules can be removed without retraining. This way, efficient modules are retained and play a more significant role in a highly weighted form.

\begin{figure*}
	\centering
     \includegraphics[width=0.9\textwidth]{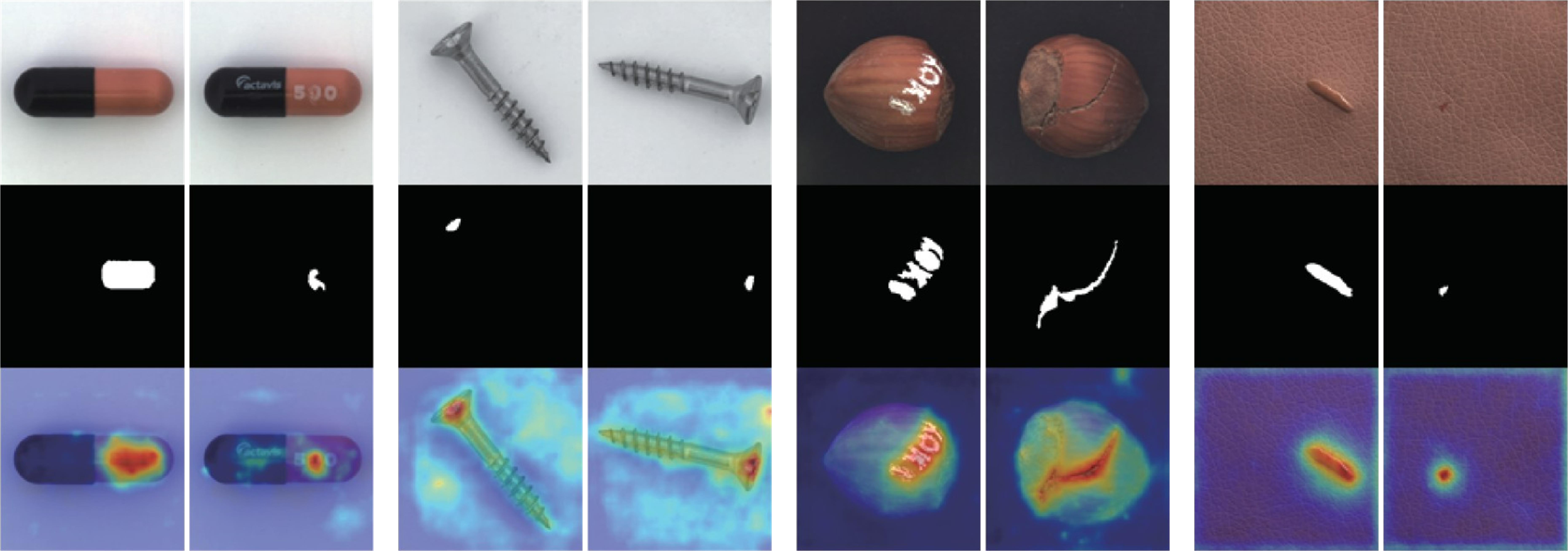}
	\caption{\centering{Multi-scale features imitate the anomaly localization effect, and each row represents the original picture, GT, and prediction probability plot}.}
	\label{fig3}
\end{figure*}

\section{Experiment}
\subsection{Experimental Details}
This paper uses a pre-trained network trained on a classification task in the ImageNet dataset as a teacher network, and the student network is randomly initialized. Use 4 feature layers to build multi-layer feature combinations, such as the output of ResNet's $conv2\_x$ to $conv5\_x$ module, and set the input image size to $[128, 256, 384]$ to build an image pyramid. The model uses the SGD optimizer with momentum set to 0.9 and the learning rate is 0.5. The batch size is set to 16, and 600 cycles of training are performed.

\subsection{Datasets And Evaluation Metrics}
This paper uses the MVTec \cite{ref27} dataset for experiments. MVTec is an industrial inspection dataset containing more than 5,000 images of 15 categories of industrial products. Each category contains normal samples for training and abnormal samples for testing. This setup directly conforms to unsupervised anomaly detection, and each class is trained and tested separately.

The experiment used threshold-independent evaluation indicators and receiver operator characteristics (ROC) and calculated the area under the curve (AUROC) as a numerical index. In addition, this paper uses another evaluation indicator, per-region overlap (PRO), which equally considers the prediction accuracy of all connected regions. Similarly, calculate its false positive rate (FPR) area under the curve (AUPRO) as a numerical indicator. Here, the area of the curve from 0\% to 30\% of the false positive rate is calculated and normalized.

\subsection{Experimental Results}
The feature mimicry network is compared to several methods of the same period. Here, the teacher network structure is selected as WideResNet-50, and the experimental results of the image pyramid multi-scale strategy are used. The experimental results of the module weight search are not compared here because the module weight search task uses a part of the data as a validation set, and its test set is not the same as other methods. The comparison methods were derived from primary sources. The experimental results are shown in Table \ref{tb1}, and the proposed method outperforms other methods in performance.

The visualization effect of the discriminant feature map is shown in Fig. \ref{fig3}, and it can be seen that this method can have a good detection effect for various types of objects and abnormal areas of different types and sizes.

\begin{table}[!t]
\renewcommand{\arraystretch}{1.3}
\caption{\label{tb1}\centering{Different Methods}}
\centering
\setlength{\tabcolsep}{1mm}
\resizebox{\linewidth}{!}
{
\begin{tabular}{ccccccc}
\hline
   &   GAVGA \cite{ref18}   &   PatchSVDD \cite{ref27}  &    STAD \cite{ref20}    &   PaDiM \cite{ref21}   &   SPADE \cite{ref22}   &   Ours \\
\hline
AUROC   &   0.93    &   0.957   &   $-$   &   0.975   &   0.965   &   \textbf{0.983} \\
AUPRO   &   $-$     &   $-$      &   0.857   &   0.921   &   0.917   &   \textbf{0.927}  \\
\hline
\end{tabular}
}
\end{table}

\begin{table}[!t]
\renewcommand{\arraystretch}{1.3}
\caption{\label{tb2}\centering{Different Multi-scale Network (AUROC)}}
\centering
\setlength{\tabcolsep}{3mm}
\resizebox{\linewidth}{!}{
\begin{tabular}{cccc}
\hline
   &   ResNet-18   &   ResNet-50   &    WideResNet-50  \\
\hline
$[2, 3, 4]$   &   0.971  &    0.977   &   0.979   \\
$[2, 3, 4, 5]$    &   0.971   &   0. 975  &   0.975   \\
\hline
\end{tabular}}
\end{table}

\subsection{Ablation Study}
\subsubsection{Backbone Network Structure And Feature Level Combination}
In this section, experiments are performed on different backbone network structures and feature layer combinations, and the input image size is fixed at 256. ResNet-18, ResNet-50, and WideResNet-50 and each structure corresponding to the use of the first three feature layers and all four feature layers are shown in Table \ref{tb2}. Experimental results show that larger models, such as WideResNet-50, outperform smaller models. Large model generation is more discriminative and better for distinguishing between learned and unlearned data. For the experimental results of different feature layer combinations, this paper finds that the first three-layer feature combinations are better than all four-layer feature combinations on all infrastructures. Combined with the deepest layer of features, the effect decreases. On the one hand, because the resolution is too small, direct upsampling leads to inaccurate results. Conversely, the semantic-level features that are too abstract are more inclined to the classification task of the pre-trained network.

\subsubsection{Multi-scale process}
Experimental results for each object class at single-input and multi-input resolution are reported here, using ResNet-50 as the base model to verify the effectiveness of the multiscale strategy. The input sizes are set to 128, 256, and 384, and the experimental results are shown in Table \ref{tb3}. These data show that the desired receptive field for each object class to generate the most suitable discriminant features is different. Small input sizes are more suitable for object classes with global size anomalies, such as transistors. Conversely, larger input sizes work better for object classes with tiny anomalies. It is more efficient to use multiple input size combinations to accommodate the anomaly detection of general categories.

\begin{table}[!t]
\renewcommand{\arraystretch}{1.3}
\caption{\label{tb3}\centering{Different Input Size Models (AUROC)}}
\centering
\setlength{\tabcolsep}{3mm}
\resizebox{\linewidth}{!}{
\begin{tabular}{ccccc}
\hline
   &   128   &   256   &    384     &   multi-scale  \\
\hline
Carpet  &   0.983   &   0.989   &   0.990   &   \textbf{0.991}   \\
Grid    &   0.948   &   0.986   &   \textbf{0.991}  &   0.989   \\
Leather &   0.993   &   0.994   &   0.992   &   \textbf{0.994}  \\
Tile    &   0.957   &   0.966   &   0.970   &   \textbf{0.970}  \\
Wood    &   0.943   &   0.953   &   \textbf{0.957}  &   0.956   \\
Bottle  &   0.980   &   0.987   &   \textbf{0.987}  &   0.986   \\
Cable   &   0.975   &   0.969   &   0.954   &   \textbf{0.975}  \\
Capsule &   0.972   &   0.987   &   0.988   &   \textbf{0.989}  \\
Hazelnut &  0.989   &   0.991   &   0.991   &   \textbf{0.992}  \\
Metalnut & 0.975   &   0.982   &   0.979   &   \textbf{0.982}  \\
Pill    &   0.969   &   0.976   &   0.974   &   \textbf{0.977}  \\
Screw   &   0.973   &   0.988   &   0.990   &    \textbf{0.971}  \\ 
Toothbrush & 0.982  &   0.988   &   \textbf{0.991}  &   0.989   \\
Transistor  &   \textbf{0.978}  &   0.948   &   0.874   &   0.968   \\
Zipper  &   0.958   &   0.984   &   \textbf{0.989}  &   0.985   \\
avg     &   0.972   &   0.979   &  0.974   &   \textbf{0.982}   \\
\hline
\end{tabular}}
\end{table}

\subsubsection{Module Weight Search Policy}
To further verify the effectiveness of the module weight search strategy based on the validation set, the anomaly data category provided by the MVTec dataset is used as a priori to experiment on the relationship between the validation set selection method and the weight search strategy. Here the selection of the validation set is divided according to the percentage of coverage of abnormal data categories (category coverage), which approximates the distribution similarity of the verification set and the test set. Extract 10\% of the data from each selected test set exception category to construct the validation set, and calculate the test accuracy under different exception category coverage. The experimental results are shown in Table \ref{tb4}, with 0\% representing no weight search. As seen from Table \ref{tb4}, the model's accuracy on the validation set decreases as the coverage of anomaly categories increases. This is because the increased number of anomalies makes it more challenging to learn the model.

Conversely, the model's accuracy on the test set increases as the coverage of anomaly categories increases. When the test set is consistent with the distribution of the validation set (100\%), the model achieves the highest accuracy. This phenomenon shows that the improvement of anomaly category coverage can effectively overcome the model's overfitting problem and improve the model's accuracy on the test set.

\begin{table}[!t]
\renewcommand{\arraystretch}{1.3}
\caption{\label{tb4}\centering{Exception Category Coverage with Different Validation Sets (AUROC)}}
\centering
\setlength{\tabcolsep}{3mm}
\resizebox{\linewidth}{!}{
\begin{tabular}{cccccc}
\hline
   &   0\%   &   25\%   &    50\%     &   75\%  &   100\%  \\
\hline
val     &       &   0.984   &   0.981   &   0.982   &   0.983   \\
test    &   0.982   &   0.984   &   0.984   &   0.985   &   0.985   \\
\hline
\end{tabular}}
\end{table}

Table \ref{tb5} shows the effect of assigning weights to student modules and the model accuracy of the modules corresponding to the maximum $k$ reserved. Weight allocation can improve the model effect from Table \ref{tb5}. Test the experimental results with $k$-values of 4, 5, and 6, where $k$-values of 6 have the highest accuracy. Overall, the $k$ modules with the highest retention weights are better than the basic model, indicating that after weight allocation and removing low-importance modules, the network can not only make the network more streamlined but also improve the accuracy performance of the network to a certain extent.

\begin{table}[!t]
\renewcommand{\arraystretch}{1.3}
\caption{\label{tb5}\centering{Weight Search and Module Retention Model Performance (AUROC)}}
\centering
\setlength{\tabcolsep}{3mm}
\resizebox{\linewidth}{!}{
\begin{tabular}{ccccc}
\hline
Weight Avg   &   Weight assignment   &   $K=4$   &     $K=5$     &    $K=6$  \\
\hline
0.983     &   0.986    &   0.985   &   0.986   &   \textbf{0.987}  \\
\hline
\end{tabular}}
\end{table}

\section{Conclusions}
This paper proposes a multi-scale feature imitation anomaly localization method based on a teacher-student network. The student network is designed as a separate group of student modules. Each module is trained and reasoned separately, combined with the pyramid of features and images to carry out multi-scale, multi-level abnormal discriminant reasoning. In addition, a module importance weight allocation strategy based on gradient descent optimization is proposed, which optimizes the network structure and further improves the network's performance. Experimental results show that the proposed method has better performance than similar methods and can effectively realize the localization detection of various types of objects and abnormal types of different sizes.

%\section*{References}


\begin{thebibliography}{00}
\bibitem{ref8}
Jinwon An and Sungzoon Cho.
\newblock Variational autoencoder based anomaly detection using reconstruction
  probability.
\newblock {\em Special Lecture on IE}, 2(1):1--18, 2015.

\bibitem{ref17}
Christoph Baur, Benedikt Wiestler, Shadi Albarqouni, and Nassir Navab.
\newblock Deep autoencoding models for unsupervised anomaly segmentation in
  brain mr images.
\newblock In {\em International MICCAI brainlesion workshop}, pages 161--169.
  Springer, 2018.

\bibitem{ref27}
Paul Bergmann, Kilian Batzner, Michael Fauser, David Sattlegger, and Carsten
  Steger.
\newblock The mvtec anomaly detection dataset: a comprehensive real-world
  dataset for unsupervised anomaly detection.
\newblock {\em International Journal of Computer Vision}, 129(4):1038--1059,
  2021.

\bibitem{ref21}
Paul Bergmann, Michael Fauser, David Sattlegger, and Carsten Steger.
\newblock Uninformed students: Student-teacher anomaly detection with
  discriminative latent embeddings.
\newblock In {\em Proceedings of the IEEE/CVF Conference on Computer Vision and
  Pattern Recognition}, pages 4183--4192, 2020.

\bibitem{ref4}
Jun~Kang Chow, Zhaoyu Su, Jimmy Wu, Pin~Siang Tan, Xin Mao, and Yu-Hsing Wang.
\newblock Anomaly detection of defects on concrete structures with the
  convolutional autoencoder.
\newblock {\em Advanced Engineering Informatics}, 45:101105, 2020.

\bibitem{ref23}
Niv Cohen and Yedid Hoshen.
\newblock Sub-image anomaly detection with deep pyramid correspondences.
\newblock {\em arXiv preprint arXiv:2005.02357}, 2020.

\bibitem{ref12}
Lucas Deecke, Robert Vandermeulen, Lukas Ruff, Stephan Mandt, and Marius Kloft.
\newblock Image anomaly detection with generative adversarial networks.
\newblock In {\em Joint european conference on machine learning and knowledge
  discovery in databases}, pages 3--17. Springer, 2018.

\bibitem{ref22}
Thomas Defard, Aleksandr Setkov, Angelique Loesch, and Romaric Audigier.
\newblock Padim: a patch distribution modeling framework for anomaly detection
  and localization.
\newblock In {\em International Conference on Pattern Recognition}, pages
  475--489. Springer, 2021.

\bibitem{ref10}
Dong Gong, Lingqiao Liu, Vuong Le, Budhaditya Saha, Moussa~Reda Mansour, Svetha
  Venkatesh, and Anton van~den Hengel.
\newblock Memorizing normality to detect anomaly: Memory-augmented deep
  autoencoder for unsupervised anomaly detection.
\newblock In {\em Proceedings of the IEEE/CVF International Conference on
  Computer Vision}, pages 1705--1714, 2019.

\bibitem{ref26}
Kaiming He, Xiangyu Zhang, Shaoqing Ren, and Jian Sun.
\newblock Deep residual learning for image recognition.
\newblock In {\em Proceedings of the IEEE conference on computer vision and
  pattern recognition}, pages 770--778, 2016.

\bibitem{ref3}
Wenqian Liu, Runze Li, Meng Zheng, Srikrishna Karanam, Ziyan Wu, Bir Bhanu,
  Richard~J Radke, and Octavia Camps.
\newblock Towards visually explaining variational autoencoders.
\newblock In {\em Proceedings of the IEEE/CVF Conference on Computer Vision and
  Pattern Recognition}, pages 8642--8651, 2020.

\bibitem{ref1}
Hui Lv, Chen Chen, Zhen Cui, Chunyan Xu, Yong Li, and Jian Yang.
\newblock Learning normal dynamics in videos with meta prototype network.
\newblock In {\em Proceedings of the IEEE/CVF Conference on Computer Vision and
  Pattern Recognition}, pages 15425--15434, 2021.

\bibitem{ref9}
Pramuditha Perera, Ramesh Nallapati, and Bing Xiang.
\newblock Ocgan: One-class novelty detection using gans with constrained latent
  representations.
\newblock In {\em Proceedings of the IEEE/CVF Conference on Computer Vision and
  Pattern Recognition}, pages 2898--2906, 2019.

\bibitem{ref15}
Pramuditha Perera and Vishal~M Patel.
\newblock Learning deep features for one-class classification.
\newblock {\em IEEE Transactions on Image Processing}, 28(11):5450--5463, 2019.

\bibitem{ref25}
Adriana Romero, Nicolas Ballas, Samira~Ebrahimi Kahou, Antoine Chassang, Carlo
  Gatta, and Yoshua Bengio.
\newblock Fitnets: Hints for thin deep nets.
\newblock {\em arXiv preprint arXiv:1412.6550}, 2014.

\bibitem{ref14}
Lukas Ruff, Robert Vandermeulen, Nico Goernitz, Lucas Deecke, Shoaib~Ahmed
  Siddiqui, Alexander Binder, Emmanuel M{\"u}ller, and Marius Kloft.
\newblock Deep one-class classification.
\newblock In {\em International conference on machine learning}, pages
  4393--4402. PMLR, 2018.

\bibitem{ref7}
Mayu Sakurada and Takehisa Yairi.
\newblock Anomaly detection using autoencoders with nonlinear dimensionality
  reduction.
\newblock In {\em Proceedings of the MLSDA 2014 2nd workshop on machine
  learning for sensory data analysis}, pages 4--11, 2014.

\bibitem{ref20}
Mohammadreza Salehi, Niousha Sadjadi, Soroosh Baselizadeh, Mohammad~H Rohban,
  and Hamid~R Rabiee.
\newblock Multiresolution knowledge distillation for anomaly detection.
\newblock In {\em Proceedings of the IEEE/CVF conference on computer vision and
  pattern recognition}, pages 14902--14912, 2021.

\bibitem{ref6}
Thomas Schlegl, Philipp Seeb{\"o}ck, Sebastian~M Waldstein, Georg Langs, and
  Ursula Schmidt-Erfurth.
\newblock f-anogan: Fast unsupervised anomaly detection with generative
  adversarial networks.
\newblock {\em Medical image analysis}, 54:30--44, 2019.

\bibitem{ref18}
Thomas Schlegl, Philipp Seeb{\"o}ck, Sebastian~M Waldstein, Ursula
  Schmidt-Erfurth, and Georg Langs.
\newblock Unsupervised anomaly detection with generative adversarial networks
  to guide marker discovery.
\newblock In {\em International conference on information processing in medical
  imaging}, pages 146--157. Springer, 2017.

\bibitem{ref13}
Bernhard Sch{\"o}lkopf, John~C Platt, John Shawe-Taylor, Alex~J Smola, and
  Robert~C Williamson.
\newblock Estimating the support of a high-dimensional distribution.
\newblock {\em Neural computation}, 13(7):1443--1471, 2001.

\bibitem{ref19}
Shashanka Venkataramanan, Kuan-Chuan Peng, Rajat~Vikram Singh, and Abhijit
  Mahalanobis.
\newblock Attention guided anomaly localization in images.
\newblock In {\em European Conference on Computer Vision}, pages 485--503.
  Springer, 2020.

\bibitem{ref24}
Guodong Wang, Shumin Han, Errui Ding, and Di Huang.
\newblock Student-teacher feature pyramid matching for unsupervised anomaly
  detection.
\newblock {\em arXiv preprint arXiv:2103.04257}, 2021.

\bibitem{ref5}
Jihun Yi and Sungroh Yoon.
\newblock Patch svdd: Patch-level svdd for anomaly detection and segmentation.
\newblock In {\em Proceedings of the Asian Conference on Computer Vision},
  2020.

\bibitem{ref16}
Jihun Yi and Sungroh Yoon.
\newblock Patch svdd: Patch-level svdd for anomaly detection and segmentation.
\newblock In {\em Proceedings of the Asian Conference on Computer Vision},
  2020.

\bibitem{ref2}
Muhammad~Zaigham Zaheer, Jin-ha Lee, Marcella Astrid, and Seung-Ik Lee.
\newblock Old is gold: Redefining the adversarially learned one-class
  classifier training paradigm.
\newblock In {\em Proceedings of the IEEE/CVF Conference on Computer Vision and
  Pattern Recognition}, pages 14183--14193, 2020.

\bibitem{ref11}
Bo Zong, Qi Song, Martin~Renqiang Min, Wei Cheng, Cristian Lumezanu, Daeki Cho,
  and Haifeng Chen.
\newblock Deep autoencoding gaussian mixture model for unsupervised anomaly
  detection.
\newblock In {\em International conference on learning representations}, 2018.
\end{thebibliography}
\end{document}